# The "Face on Mars": a photographic approach for the search of signs of past civilizations from a macroscopic point of view, factoring long-term erosion in image reconstruction

by

Vassilis S. Vassiliadis, *Ph.D.*,
*Senior Lecturer,*

Department of Chemical Engineering,
University of Cambridge,
Pembroke Street,
Cambridge CB2 3RA, United Kingdom.

15 December 2005

## Abstract

This short article presents an alternative view of high resolution imaging from various sources with the aim of the discovery of potential sites of archaeological importance, or sites that exhibit "anomalies" such that they may merit closer inspection and analysis. Use is made of the infamous photos of Cydonia on Mars, and in particular of the site that became anecdotally known as the "face on Mars", following the 1976 Viking photos under low resolution. It is conjectured, and to a certain extent demonstrated here, that it is possible for advanced civilizations to factor in erosion by natural processes into a large scale design so that main features be preserved even with the passage of millions of years. Alternatively viewed, even without such intent embedded in a design left for posterity, it is possible that a gigantic construction may naturally decay in such a way that even cataclysmic (massive) events may leave sufficient information intact with the passage of time, provided one changes the point of view from high resolution images to enhanced blurred renderings of the sites in question. If one takes the intentional design for erosion to be taken into account, then it is also possible to conjecture that it is not necessary for the original structure ("monument") to be designed such that it resembles something if viewed at high resolution in the first place; in fact it is possible to treat elements in the environment as large scale "pixels", as natural monolithic blocks, so that when viewed from a distance under slightly unfocused conditions they produce the desired effect, thus lasting for millennia. In terms as to how such an enormous project can be accomplished one does not need to think in terms of detailed painstaking work, but rather once a design is conceived then even large scale catastrophic "sculpting" (large scale explosions) can achieve the desired effect. In this work, it is shown that the photo of 1976 by Viking, showing an impression of an "alien" face when viewed with some imagination, can be reconstructed to yield the same result if a modern high resolution image (2001, Global Mars Surveyor) is blurred and is edge-enhanced. This is demonstrated by using standard modern digital image manipulation software in the present article.

## 1. The original photos of the Viking mission in 1976

A detail from a site on Mars revealed the following images, found freely on the internet:

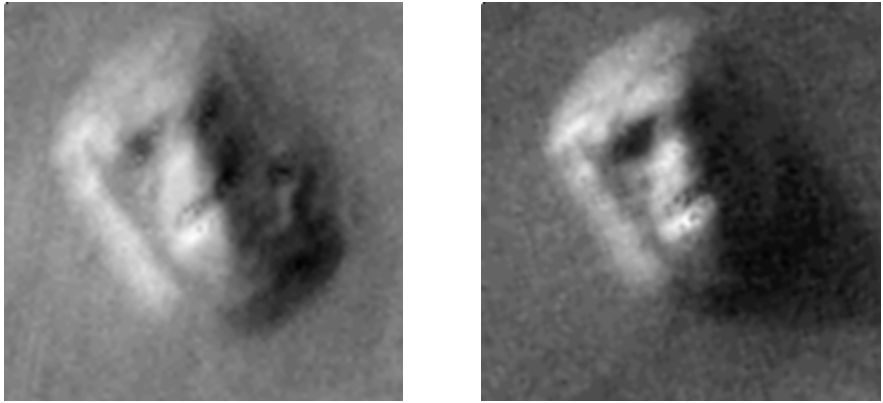

The two images are of the identical structure, the one on the right hand side under more contrasty conditions. These images have caused a stir over the years, being interpreted by a large number of people as indications of a monument left by ancient alien civilizations on the now barren planet. The overall area where the structure is found is called Cydonia, and two pictures of it is shown below.

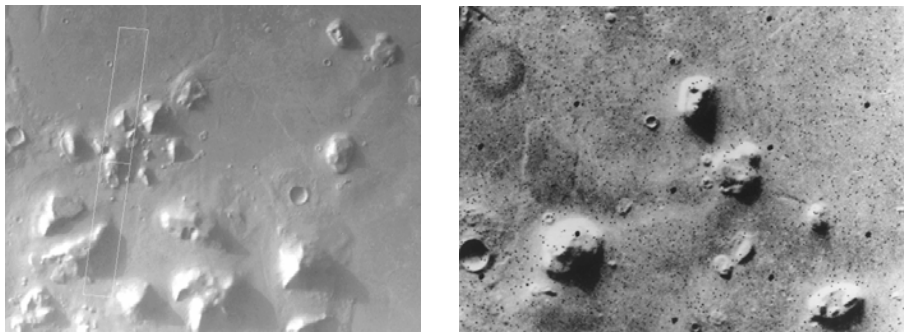

Many people, again by looking at the site pictures, were quick to discover symmetries, pyramidal structures and the like in them. Some have also done painstaking work so as to use advanced image processing techniques and algorithms to demonstrate how the sites would look under different angles of illumination, projections etc., in trying to demonstrate that there is purpose and design in them.

The point of view of this work is not to analyse the validity of such claims, neither to prove nor to disprove them. However, motivated by observations by several serious people it was decided to adopt a different point of view to demonstrate some potential principles by working with the modern high resolution images of the "face".

## 2. Blurring images in modern image enhancement

Several modern techniques use blurred versions of original images to either enhance edges or to achieve what is called local contrast enhancement (LCE), either manually or "automatically". People familiar with software such as Adobe Photoshop or GIMP use the feature of the "unsharp mask" to achieve this effect.

Here we are proposing the use of blurred images directly in order to "face-lift" large scale structures that may have been deliberately constructed, so as to reveal intent in their design or identify regions of potential archaeological significance. A combination of edge detection and LCE over the blurred image (derived from a sharp high resolution original) can further enhance features that may be artificial. Signs of such designs may be unnatural curvatures, angles and symmetries. Archaelogists in general have decided to dig based on much less information until they made a discovery, so it is envisaged that the proposed methodology might have some usefulness for the evaluation of new sites as well as the re-evaluation of older sites where large structures dominate.

## 3. High resolution image of the "Face on Mars" and blurred versions

The first steps in this work is to blur at increasing levels a relatively recent image of the "face". In itself the image is totally uninteresting as it is evident that the region apart from dominant oval shape it contains just "rubble" scattered around by natural processes. The image we start from is the following:

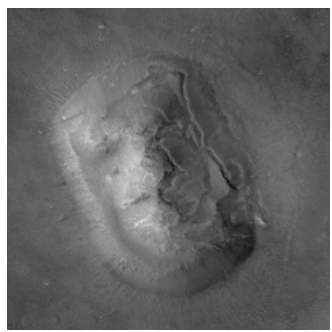

A series of progressively blurred images is shown in the table below. To blur the original image a Gaussian blur filter was used with GIMP, measuring how many pixels were used to "diffuse" tonality around each original pixel.

| **Blur level (Gaussian blur in pixels)** | **Image** |
|---|---|
| Original image – blur : 0 | 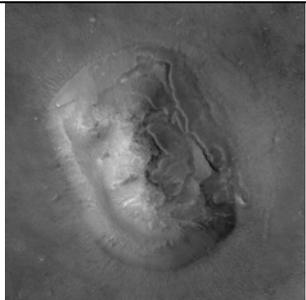 |
| blur : 10 | 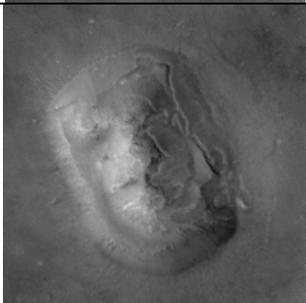 |
| blur : 20 | 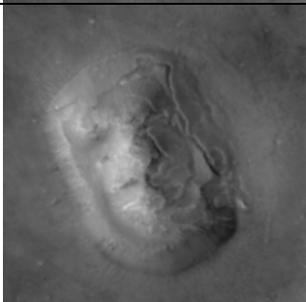 |
| blur : 50 | 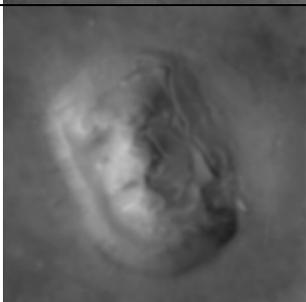 |
| blur : 100 | 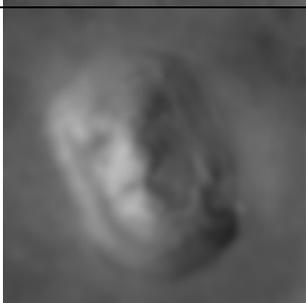 |

*Unpublished paper*      *Vassilios S. Vassiliadis*

| | | | |
|---|---|---|---|
| blur : 150 | | 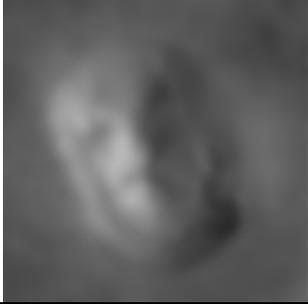 | |
| blur : 200 | | 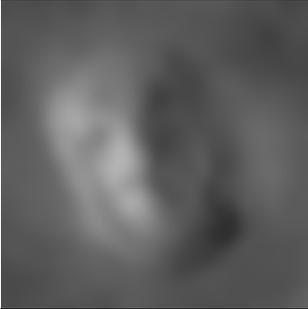 | |

From the table above, the blur level 200 is rather excessive, while the one at 150 is deemed good enough for our demonstration. Futher work on that image is shown in the following sequence of images in the table below.

| *Description* | *Image* |
|---|---|
| Enhanced contrast and tonality level adjustments | 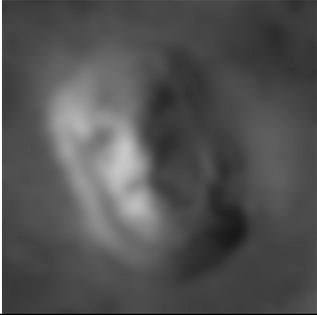 |
| Edges of the above image. GIMP setting was Sobel with intense rendering options. Also the edge image was inverted and contrast/tonality adjustments were made to intensify the edges. | 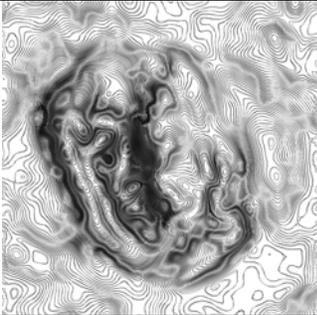 |
| A composite of the two above images was effected in GIMP using an overlay blending mode with the edges blended at 9.0% opacity to the top image of the "face". | 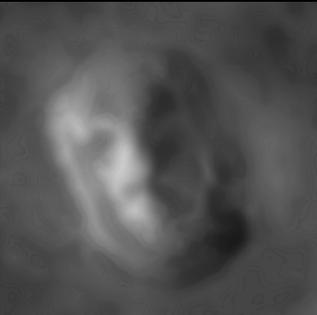 |

The final image is displayed in larger size below, followed by a copy of one of the original Viking images blown up to the same size for comparison.

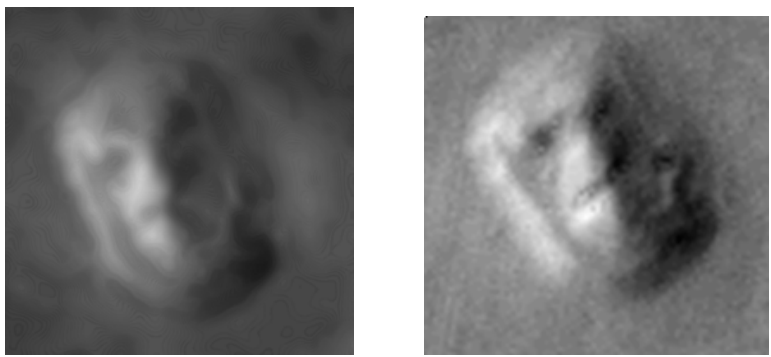

When the image is viewed from some distance it seems that there are some symmetries imbedded in it: an almost dominant left-right (tilted diagonally) symmetry with some breakage at the lower right hand side end of the structure. As to its significance, there will be no comment in this work. However, there will be some discussion to follow from the principles demonstrated here.

Before closing this section we discuss a potential strategy of how someone could go about to make a "face" on Mars, hypothetically speaking. The following table outlines the strategy as a step by step procedure, the final stage of which is the viewer blurring out the detail to perceive the underlying intentional design.

| *Description* | *Image* |
|---|---|
| Topographic examination of site. The choice is to find something mountainous with a peak that could serve as a lower part (mouth/nose) of a "sculpture". <br> The lines drawn in are the key defining contour lines of assumed equal height. The middle circle encompasses the highest point in the natural formation. | 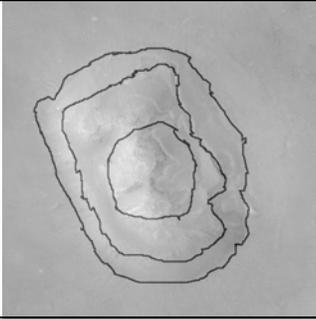 |
| The only artificial modifications required are to bring in rocks in the place of the eyes, either add them in or dig the ground out to create curvature that will create shadows from any point of illumination. | 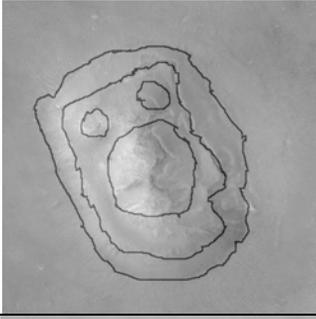 |
| Because of the mountainous structure of the formation, any angle of illumination will produce shadows going in the opposite direction from the source of light. In other words it does not matter how light falls: one will always see a "face" if that is what they choose to perceive. This would be the high resolution image that has really no information if viewed in great detail. | 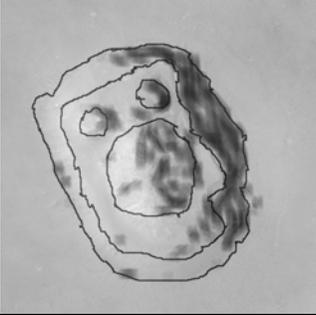 |
| The viewer adds their blurring in to see what the underlying structure could signify. A contrast enhancement on the blurred image at 150 pixels Gaussian blur was also performed to enhance the potential intent of the "artist". | 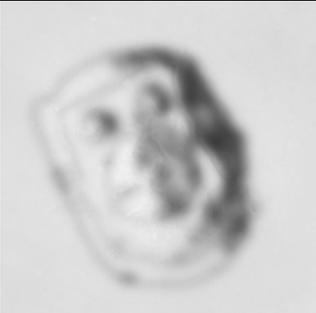 |

## 4. Discussion and conclusions

In this work we demonstrate that it is possible to use blurring techniques in order to reveal large scale features of very old structures, whether by artificial design or natural processes.

The demonstration using the "Face on Mars" was more in terms of carrying out an experiment in an exotic locale so as to demonstrate the value of this technique. Further to this, it is possible to use such technology to design landmarks that are to be left for posterity, with the intent that they be revealed when appropriate viewing techniques are applied.

For example, in our archaeology we have examples such as the large structures of the pyramids. Despite years of defacing, intentional e.g. to strip of the marble (?) outer shell, or by sand and wind erosion, their shape is still clearly defined especially if viewed from a distance where irregularities "wash out". Similarly the Sphinx at Giza shows what happens by erosion: the actual details diminish as time goes by, whereas the overall shape of a "lion" is still retained macroscopically. By counterexample, the overly detailed designs of Greek and Roman temples seem to not age as well as the large constructions of the Egyptians.

This is also more interesting given that the sandy environment of Egypt is more "corrosive" than say modern pollution in Athens or Rome! Precisely because of the large bulk and pre-existing curvature in the design the overall impression of Egyptian architectural works is still retained almost intact perceptually, despite their being more than twice the age and in harsher conditions.

From a futuristic point of view, it would be interesting to assume that an advanced civilization would understand that if its mark were to endure millennia, then the only available resort for such guarantee would be to use two things from the start:

  a) A large scale structure, in large part natural if possible (saves time and effort)
  b) Use of "pixelization" principles, as demonstrated in this paper, so as to ensure that added features form easy-to-add elements without much detailed effort

Adopting an even more "romantic" or science fiction point of view, one could say that such an advanced civilization would resort to something as simple as that, which yet contains a "catch": the challenge for another intelligent species in a nearby planet to choose not to see in high resolution, but to blur an image of a site so that they see their message in the sand across millions of years…